\documentclass{article}

\PassOptionsToPackage{numbers, compress}{natbib}

\usepackage{booktabs}
\usepackage{natbib}

\usepackage[preprint]{neurips_2023}

\usepackage[utf8]{inputenc} 
\usepackage[T1]{fontenc}    
\usepackage{hyperref}       
\usepackage{url}            
\usepackage{booktabs}       
\usepackage{amsfonts}       
\usepackage{nicefrac}       
\usepackage{microtype}      
\usepackage{xcolor}         
\usepackage{subcaption}   
\usepackage{graphicx}
\usepackage{algorithm}
\usepackage{amsmath}
\usepackage{todonotes}
\usepackage{algorithmic}
\title{Reviving Shift Equivariance in Vision Transformers}
\usepackage{amsthm}
\usepackage{float}
\usepackage{amssymb}
\usepackage{enumitem}

\newtheorem{corollary}{Corollary}[section]
\newtheorem{lemma}{Lemma}[section] 

\usepackage{soul}

\author{%
  Peijian Ding \\
  University of Maryland\\
  \texttt{pding@umd.edu} \\
  \And
  Davit Soselia \\
  University of Maryland\\
  \texttt{dsoselia@cs.umd.edu} \\
  \And
  Thomas Armstrong \\
  University of Maryland\\
  \texttt{tarmst@terpmail.umd.edu} \\
  \And
  Jiahao Su \\
  University of Maryland \\
  \texttt{jiahaosu@umd.edu} \\
  \And
  Furong Huang \\
  University of Maryland\\
  \texttt{furongh@umd.edu} \\
}
\begin{document}

\maketitle

\begin{abstract}
Shift equivariance is a fundamental principle that governs how we perceive the world — our recognition of an object remains invariant with respect to shifts. Transformers have gained immense popularity due to their effectiveness in both language and vision tasks. While the self-attention operator in vision transformers (ViT) is permutation-equivariant and thus shift-equivariant, patch embedding, positional encoding, and subsampled attention in ViT variants can disrupt this property, resulting in inconsistent predictions even under small shift perturbations. Although there is a growing trend in incorporating the inductive bias of convolutional neural networks (CNNs) into vision transformers, it does not fully address the issue. We propose an adaptive polyphase anchoring algorithm that can be seamlessly integrated into vision transformer models to ensure shift-equivariance in patch embedding and subsampled attention modules, such as window attention and global subsampled attention. Furthermore, we utilize depth-wise convolution to encode positional information. Our algorithms enable ViT, and its variants such as Twins to achieve 100\% consistency with respect to input shift, demonstrate robustness to cropping, flipping, and affine transformations, and maintain consistent predictions even when the original models lose 20 percentage points on average when shifted by just a few pixels with Twins' accuracy dropping from 80.57\% to 62.40\%.
\end{abstract}


\section{Introduction}
Inductive bias refers to a set of assumptions or beliefs that guide the design of machine learning algorithms, aiming to reduce the search space for an optimal model. Humans possess the remarkable ability to perceive and recognize objects despite variations such as deformations, occlusions, and translations. The success of convolutional neural networks (CNNs) can be attributed to their inductive bias of shift equivariance, which allows them to mimic this human ability. For instance, when we observe an image of a cat, shifting the cat's position within the image does not affect our recognition of the cat, and we remain aware of the shift that has occurred. \par

Transformers have emerged as strong alternatives to CNNs in computer vision, and their success in natural language processing further highlights their potential as a research area. A key element of transformers is the self-attention module, which exhibits permutation-equivariance. However, despite this strong inductive bias, vision transformers are neither shift-equivariant nor permutation-equivariant. This is due to the patch embedding, positional embedding, and subsampled attention (in the case of ViT variants), which disrupt shift equivariance as input shifts lead to different pixels within each image patch or different tokens in each window, resulting in different computations.\par

Integrating CNNs with vision transformers can partially address the lack of shift equivariance, but it does not fully resolve the issue. The original design of vision transformers already incorporates convolution within its architecture; the patch embedding layer is functionally equivalent to strided convolution, but its downsampling operation disrupts shift equivariance. Although CoAtNet proposes a relative attention method that combines depthwise convolution with attention to achieve shift equivariance, this approach still requires computing full global attention, and shift equivariance is not maintained when downsampled attention is required for computational efficiency \citep{twins, maxvit, davit}. Both MaxViT \citep{maxvit} and Twins transformer \citep{twins} utilize depth-wise convolution to encode positional information, but their block attention (or window attention) and strided convolution are not shift equivariant. 

In this work, we seek to explicitly incorporate the inductive bias of CNNs into vision transformers, fully addressing the issue of shift equivariance. We achieve this by introducing replaceable modules that enable a wide range of vision transformer models to be shift-equivariant \citep{vit, cswin, twins, maxvit, conditional, pvt, longformer}. Our proposed solution is a nonlinear operator --- the polyphase anchoring algorithm. By consistently selecting the maximum polyphase as anchors for computing strided convolution and subsampled attentions, we ensure shift equivariance. Furthermore, we employ depthwise convolution with circular padding to encode positional information, as opposed to using the absolute positional embedding in \citet{vit} or the relative positional embedding in \citet{swin, swinv2}, which are not shift equivariant. \par

\begin{figure}[htbp]
         \centering
         \includegraphics[width=\textwidth]{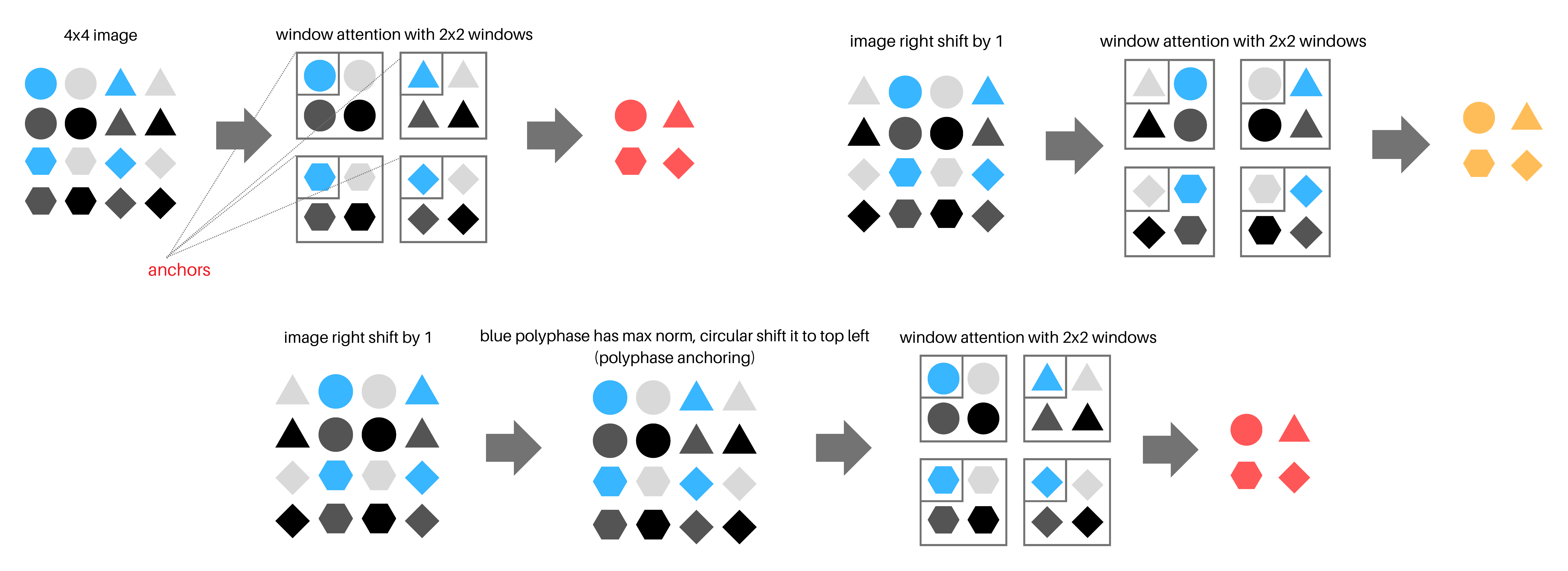}
       \caption{The maximum polyphase is colored in blue. Each shape with distinct color represents a token. We also illustrate the concept of \emph{anchors} --- the top left coordinate in each window.  The red and yellow shapes indicate that window attention produces inconsistent predictions on shifted image whereas the composition of polyphase anchoring and window attention does not. }
        \label{fig:poly}
\end{figure}

Our primary contributions in this work are as follows:
\begin{itemize}
    \item We present versatile and adaptable modules that seamlessly integrate with vision transformer models, leading to improved performance in vision transformer models.
    \item Our innovative approach incorporates an input-adaptive nonlinear operator that guarantees shift-equivariance, resulting in more robust and reliable outcomes across a range of visual tasks.
    \item We equip vision transformers with complete shift-equivariance capabilities, substantiating our approach with both theoretical geometric guarantees and empirical evidence that underscore its effectiveness and practical applicability.
\end{itemize}

\section{Preliminaries}
To effectively restore shift-equivariance in vision transformers, it is necessary to examine each individual module and identify the components responsible for disrupting the shift-equivariance property. To achieve this, we must first establish formal definitions for equivariance and self-attention. 

\subsection{Equivariance}
Equivariance serves as a formal concept of consistency under transformations \citep{lie}. A function $f : V_1 \to V_2$ is considered equivariant to transformations from a symmetry group $G$ if applying the symmetry to the input of $f$ produces the same result as applying it to the output:
\begin{align}
\forall g \in G : f(g \cdot x) = g' \cdot f(x)
\end{align}

Here, $\cdot$ denotes the linear mapping of the input by the representation of group elements in $G$. Throughout this paper, all instances of $\cdot$ adhere to this definition. When $g = g'$, the function is referred to as G-equivariant. If $g'$ is the identity, the function is G-invariant. For cases where $g \neq g'$, the function is considered generally equivariant. General equivariance is a valuable concept when the input and output spaces have different dimensions. When $G$ represents the translation group, the above definition yields shift-equivariance. \par

\subsection{Self-attention}
\label{sec:attn}
A self-attention operator $A_s$ exhibits permutation-equivariance. Let $X$ represent the input matrix, and $T_{\pi}$ denote any spatial permutation. We can express this as:
\begin{align}
A_s(T_\pi(X)) = T_\pi(A_s (X)).
\end{align}

$A_s$ is the self-attention operator with parameter matrices $W_q \in \mathbb{R}^{d\times d_k}$, $W_k \in \mathbb{R}^{d\times d_k}$, and $W_v \in \mathbb{R}^{d\times d_v}$:
\begin{align}
A_s = \text{SoftMax}(XW_q (XW_k)^T)XW_v = \text{SoftMax}(QK^T )V.
\end{align}

\section{Approach}
To achieve model-wise shift-equivariance, we first detect the modules that lack shift equivariance. We then introduce a polyphase anchoring algorithm to ensure shift-equivariance for strided convolution, window attention, and global subsampled attention. Finally, we use depthwise convolution with circular padding to guarantee shift-equivariance in positional encoding. As the composition of shift-equivariant functions remains shift-equivariant, we obtain a shift-equivariant model. 

\subsection{Detecting modules lacking shift equivariance}
\label{sec: detect}
Vision transformers consist of a patch embedding layer, positional encoding, transformer blocks, and MLP layers. We analyze each of these modules in ViT and its variants, discovering the following:
\begin{itemize}[leftmargin=20pt]
\item Patch embedding layer (strided convolution) is not shift-equivariant due to downsampling.
\item Absolute positional encoding \citep{vit} and relative positional embedding \citep{swin, swinv2} are not shift-equivariant.
\item Normalization, global self-attention, and MLP layers are shift-equivariant.
\item Subsampled attentions such as window attention \citep{swin, swinv2, maxvit, twins} and global subsampled attention \citep{twins} are not shift-equivariant.
\end{itemize}

\paragraph{Patch embedding} converts image patches into sequence vector representations through strided convolution. However, strided convolution is not shift-equivariant due to downsampling, as addressed by \citet{blurpool, poly-d}. Figures \ref{fig:poly} illustrate that the image patching layer, or strided convolution, is not shift-equivariant. When an image is shifted by a pixel, the pixels in each window change, leading to different computations.

\begin{figure}
    \centering
    \includegraphics[width=\textwidth]{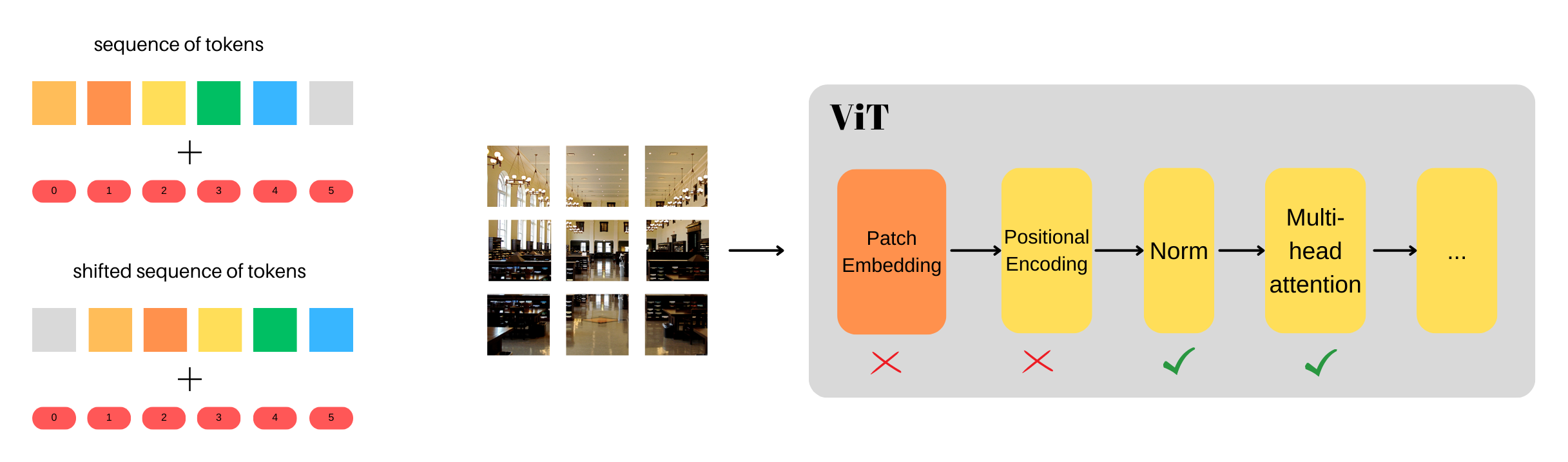}
    \caption{The left figure illustrates that the identical absolute positional encoding is applied to both the input and its circularly shifted counterpart, resulting in a lack of shift-equivariance. The right figure highlights that, in the context of ViT, patch embedding and positional encoding do not exhibit shift-equivariant properties.}
    \label{fig:pos}
\end{figure}

\paragraph{Positional encoding} is a method for incorporating spatial location information into tokens, the representations of input image patches. Popular positional encoding techniques such as absolute positional encoding in ViT \citep{vit} and relative positional encoding in Swin \citep{swin, swinv2} are not shift-equivariant. Absolute positional encoding \citep{vit} adds the absolute positional information to input tokens by considering an input image as a sequence or a grid of patches \citep{vit, pvt}. Trivially, absolute positional embedding is not shift-equivariant because the same absolute positional information is added to the input tokens regardless of shift, as shown in Figure \ref{fig:pos}
 The relative positional embedding introduced by \citet{swin} is the following: 
\begin{equation}
    \text{Attention}(Q, K, V ) = \text{SoftMax}(\frac{QK^T}{\sqrt{d}}+B )V,
\end{equation}
where $B\in R^{M^2\times M^2}$ is the relative position bias term for each head; $Q, K, V \in R^{M^2\times d}$ are the query, key and value matrices; $d$ is the query/key dimension, and $M^2$ is the number of patches in a window. Although self-attention is permutation equivariant, self-attention with relative position bias is not shift-equivariant. Further mathematical deductions are provided in the Appendix.

\paragraph{Normalization layers} standardize the input data or activations of preceding layers to stabilize training and enhance model performance by ensuring consistent scales and distributions. Both batch normalization and layer normalization are shift-equivariant. Trivially, normalizing a shifted input along batch and feature dimensions is equivalent to shifting the normalized input. \par

\textbf{MLP layers}, or Multi-Layer Perceptron layers, are a sequence of feedforward neural network layers that perform a linear transformation followed by a non-linear activation function. An MLP layer is shift-equivariant. In layer $l$ of an MLP model, we have:
\begin{align}
    h^{(l)} = \phi(xW^{(l)} + b^{(l)})
\end{align}
where $x$ is a row vector, $W$ is a weight matrix, and $b$ is a bias term. Given an input matrix $X$ whose row vectors are tokens, it is obvious that the MLP layer is shift equivariant with respect to input tokens. \par 

In the ViT architecture, we have identified that MLP layers and normalization layers are shift-equivariant, while patch embedding and positional encoding are not. ViT variants \citep{swin, swinv2, conditional, cswin, pvt, maxvit} introduce additional challenges for shift-equivariance, as they typically employ subsampled attention operations to reduce the quadratic computational complexity with respect to the number of tokens in global self-attention.\par  

s
\paragraph{Subsampled attentions} are streamlined versions of global self-attentions that can be classified into two categories: local and global \citep{swin, swinv2, twins, longformer, maxvit}. Local attention is typically employed in conjunction with subsampled global attention to encode substantial spatial information while avoiding excessive computational costs \cite{maxvit, twins, conditional, longformer}. However, the use of subsampled attentions often results in a lack of shift-equivariance due to downsampling. Consequently, addressing the shift-equivariance issue in these subsampled attentions is crucial.

The most prevalent local attention mechanism is window attention, while a popular subsampled global attention variant is the global subsampled self-attention (GSA) introduced by \citet{twins}. To directly tackle the lack of shift-equivariance, we propose the polyphase anchoring algorithm as a solution. This approach is designed to maintain spatial information while reducing computational complexity, thus promoting shift-equivariance in subsampled attention mechanisms.\par

\subsection{Polyphase anchoring algorithm}

\subsubsection{Introduction}
Inspired by the concept of adaptive polyphase sampling presented in \citet{poly-d}, we propose the polyphase anchoring algorithm, an efficient technique that can be seamlessly integrated with various types of subsampled attention operators \citep{maxvit, twins, swin, swinv2, cswin, davit} to ensure shift-equivariance. The algorithm is implemented as an autograd function in PyTorch, making it simple to incorporate into deep learning models.\par 

Polyphase anchoring identifies the maximum $L_p$ norm polyphase and shifts the input accordingly, so that the maximum polyphase aligns with the anchor positions of window attention, as illustrated in Figure \ref{fig:poly}. \emph{Anchors} of window attention represent the set of coordinates at the top-left of each window, as depicted in Figure \ref{fig:poly}.
\begin{align}
\text{anchors} = \{(i,j) \mid i\equiv0\pmod{s}, \quad j\equiv 0 \pmod{s}, \quad i \leq H, j \leq W, i,j \in \mathbb{Z}\}.
\end{align}
Here, $s\times s$ denotes the size of the window in window attention, $(i,j)$ is a coordinate on a 2D grid.

 Algorithm \ref{alg:poly_wa} demonstrates polyphase ordering. For the brevity in Algorithm \ref{alg:poly_wa}, we define the polyphase $X_{pq}$ mathematically here. Let $S_{pq}$ be some polyphase whose first element is at coordinate $(p,q)$ on the 2D grid, $S_{pq} = \{(i+p, j+q)  \mid i \equiv 0 \pmod{s},\quad  j \equiv 0 \pmod{s} \quad  i \leq H, j\leq W, i,j \in \mathbb{Z} \} $. Let $X_{pq}$ be all the tokens on the polyphase $S_{pq}$. $ X_{pq} = \{x = f(m,n)\in \mathbb{R}^C \mid  \forall (m,n) \in S_{pq} \}$, where $(m,n)$ is a coordinate of some token in $S_{pq}$ and  $f$ is a mapping from a coordinate to its corresponding token. 

\begin{algorithm}
\caption{Polyphase anchoring}
\label{alg:poly_wa}
\begin{algorithmic}[1]
\STATE \textbf{Input} $X \in R^{ ... \times H \times W}$, stride size $s \in \mathbb{Z}$  
\STATE Maximum polyphase is $\hat{X}_{pq} = \arg\max_{X_{pq} \mid p,q \in \{0, \dots, s-1\}} ||X_{pq}||$
\STATE $\hat{X} = g_{pq}\cdot (X)$ where $g_{pq}$ circularly shifts $X$  by $(-p, -q)$ along the last two dimensions. 
\STATE \textbf{Output}: $\hat{X}$
\end{algorithmic}
\end{algorithm}
The polyphase anchoring algorithm is a nonlinear operator that conditionally shifts the input based on its maximum $L_p$ norm polyphase. This guarantees shift-equivariance in strided convolution, subsampled attention like window attention, and GSA. As a result, the lack of shift-equivariance in patch embedding modules and subsampled attention modules in ViT variants, such as Twins \citep{twins}, is addressed.

\begin{lemma}
\label{lemma:shift}
    Polyphase anchoring operator $P$ is general equivariant with respect to $\forall g\in G$, where G is the symmetry group of translations, and $P: V \to V$ is a nonlinear operator that conditionally shift the input $X$. $\forall g\in G$, $\exists g' \in G$ s.t:
    \begin{equation}
    P(g \cdot X) = g' \cdot P(X)
    \end{equation}
\end{lemma}
\begin{corollary}
\label{coro}
$P(g\cdot X) = g'\cdot P(X)$ where $g'$ translates $ P(X)$ by an integer multiple of stride size $s$. Stride size is the distance between two consecutive tokens in the same polyphase on a 2D grid. 
\end{corollary}

\subsubsection{Addressing Shift Equivariance in Patch Embedding and Subsampled Attention}

As discussed in Section \ref{sec: detect}, strided convolution and subsampled attention methods, such as window attention and GSA, inherently lack shift equivariance. To tackle this issue, we employ the polyphase anchoring algorithm, which conditionally shifts the input based on the maximum polyphase, ensuring shift-equivariance in these modules.

Utilizing Lemma \ref{lemma:shift}, which establishes that polyphase anchoring is generally shift-equivariant, i.e., $P(g\cdot X) = g'\cdot P(X)$, and Corollary \ref{coro}, which states that $g'$ translates $P(X)$ by an integer multiple of stride size $s$,  \emph{we show that the composition of polyphase anchoring with strided convolution, window attention, and global subsampled attention results in shift-equivariant operations}. Consequently, we effectively address the lack of shift equivariance in patch embedding modules and subsampled attention modules for ViT variants such as Twins \citep{twins}. Detailed proofs are provided in Appendix.



\begin{lemma}
      Let $P$ be the polyphase anchoring operator and $\ast_s$ represent the strided convolution operator. $\forall g \in G$, where $G$ is the translation group, and $\forall s_1, s_2 \in \mathbb{Z}$ s.t $s_1 = s_2$, where $s_1$ is the stride size in the polyphase  and $s_2$ is the stride size of convolution, the composition of strided convolution and polyphase anchoring is general shift-equivariant:
     \begin{align}
         P(g\cdot X ) \ast_s h =  g' \cdot (P( X ) \ast_s h)
     \end{align}
\end{lemma}
 Here, $X$ denotes the input signal, $h$ is the convolution filter, and $\cdot$ signifies the linear mapping of the input by the representation of group elements in $G$. Furthermore, $P: V \to V$ is a nonlinear operator acting on the input space $V$.


\begin{lemma}
      Given a window attention operator $A_w$ and polyphase anchoring operator $P$, the composition of these operators is general shift-equivariant $\forall g \in G$, where $G$ is the translation group, and for $s, w \in \mathbb{Z}$ such that $s = w$, where $s$ is the stride size in the polyphase and $w \times w$ is the window size. This can be expressed as:
     \begin{equation}A_w (P(g\cdot X)) = g' \cdot A_w (P(X))\end{equation}
\end{lemma}
In the window attention operator $A_w$, we have:
\begin{equation}
A_w(X) = 
\left [ \begin{array}{ccc}
    A_s(X_{00})  & \cdots & A_s(X_{0n}) \\
    \vdots &  \vdots & \vdots \\
    A_s(X_{m0}) & \cdots  & A_s(X_{mn}),
\end{array} \right ]
\end{equation}
where $A_s$ is the self-attention operator defined in section \ref{sec:attn}, and $X_{ij}$ contains all the tokens in a window.


\begin{lemma}
    For a global subsampled attention operator $A_g$ \citep{twins} combined with a polyphase anchoring operator $P$, general shift-equivariance is achieved for $\forall g \in G$, where $G$ is the translation group, and for $s_1, s_2 \in \mathbb{Z}$ such that $s_1 = s_2$. Here, $s_1$ is the stride size in the polyphase, and $s_2$ is the stride size in the global subsampled attention. This can be expressed as:
     \begin{equation}
     A_g (P(g\cdot X)) = g' \cdot A_g (P(X))
     \end{equation}
\end{lemma}
    
In global subsampled attention, we have: \begin{equation} 
A_g(X) = \text{SoftMax}(Q K_{s}^T)V_{s} ,
\end{equation}
where $K_s$ and $V_s$ are subsampled from the full keys $K$ and values $V$ using strided convolution. \par

\subsection{Positional Embedding}

Positional embedding is another factor that can cause models to lose shift-equivariance. As demonstrated previously, both absolute positional embedding \citep{vit} and relative positional embedding \citep{swin, swinv2} are not shift-equivariant. However, the conditional positional embedding introduced by \citet{conditional} promotes shift-equivariance by employing zero-padded depthwise convolution to encode positional information. By replacing positional encoding with circularly-padded depthwise convolution, we achieve shift-equivariance. \par 
Formally, given an input tensor $\mathbf{X}$ of shape $(C_{\text{in}}, H_{\text{in}}, W_{\text{in}})$ and a set of depthwise filters $\mathbf{W} = \{W_1, W_2, \ldots, W_{C_{\text{in}}}\}$ of size $(C_{\text{in}}, k, k)$, where $C_{\text{in}}$ denotes the number of input channels and $k$ represents the kernel size, the depthwise convolution operation can be defined as:

\begin{equation}
\mathbf{Y}_i = W_i \odot \mathbf{X}_i, \quad \forall i \in \{1, 2, \ldots, C_{\text{in}}\} 
\end{equation}

Here, $\mathbf{X}_i$ denotes the $i$-th input channel, $\mathbf{Y}_i$ represents the corresponding output channel, and $\odot$ denotes the standard convolution operation. Assuming circular padding, convolution at each channel is shift-equivariant, making depthwise convolution shift-equivariant as well. Since CNNs utilize convolution to encode positional information, it is plausible that transformers can also employ depthwise convolution to encode information, as demonstrated in \citep{coatnet, conditional}. 

By ensuring shift-equivariance in patch embedding, positional embedding, and subsampled attention (in the case of ViT variants), we can achieve a truly shift-equivariant model. Furthermore, by incorporating a shift-invariant pooling operation in the classification head, we can obtain a truly shift-invariant model. \par

\section{Experiments}
In this section, we demonstrate that we can construct a $100\%$ truly shift-equivariant ViT using adaptive polyphase anchoring and positional embedding. Additionally, we show that we can make the Twins transformer \citep{twins} truly shift-equivariant, a vision transformer variant that incorporates two types of attention modules---window attention and global subsampled self-attention. Models employing our algorithm exhibit superior accuracy in fair comparisons, improved robustness under shifting, cropping, flipping, and random patch erasing, 22.4\% relative percentage point gain (or 41.6\% increase) from ViT small under worst-of-30 shift attack, and 100\% consistency under shift attacks.  \par 

\subsection{Classification Accuracy and Consistency on ImageNet-1k}
\label{sec:class}
\paragraph{Settings.} We evaluate six architectures, including ViT base, ViT small, Twins \citep{twins} base, and their shift-equivariant counterparts using polyphase anchoring on ImageNet-1k, which contains 1.28M training images and 50K validation images from 1,000 classes. For simplicity, we refer to ViT and Twins using polyphase anchoring and circular depthwise convolution as ViT-poly and Twins-poly. There are two types of Twins models introduced by \citet{twins}, and we default Twins to Twins-svt because it demonstrates superior performance.

\paragraph{Training.} To ensure fair comparisons, we conduct a controlled experiment by training each model and its polyphase counterpart with the same hyperparameters from scratch on ImageNet-1k. All training instances involve using the same data augmentation strategy.

\paragraph{Evaluation.} We measure performance using accuracy, consistency, and accuracy under small random shift from 0 to 15 pixels. Consistency \citep{poly-d} measure the likelihood of the model assigning the image and its shifted copy to the same class. 

\paragraph{Results.} Table \ref{tab:experiment_results} shows the comparison of ViT and Twins transformer models against its shift equivariant counterpart using models training from scratch with exactly the same hyperparameters. Poly version models demonstrate comparable raw accuracy, superior accuracy under random shifts, and 100\% consistency. \par

\begin{table}[htbp]
  \centering
  \caption{ImageNet1K training from scratch}
    \vspace{0.8em} 
  \label{tab:experiment_results}
  \begin{tabular}{ccccccc}
    \toprule
    \textbf{Model} & \textbf{image size} & \textbf{\#param.} & \textbf{Epochs} &\textbf{Acc. IN1K} & \textbf{Consis.}  & \textbf{Acc. Rand. S} \\ 
    \midrule
    ViT\_S & $224^2$& 22M & 300 & 75.52 & 86.61 & 74.98  \\ 
    ViT\_S-poly & $224^2$&  22M  & 300 &  \textbf{76.37} &  \textbf{100}  & \textbf{76.37}  \\  
    \midrule
    ViT\_B & $224^2$& 87M & 300 & 73.85 & 85.60  & 73.01  \\ 
    ViT\_B-poly & $224^2$& 86M  & 300 & \textbf{74.62} & \textbf{100}  & \textbf{74.62} \\  
     \midrule
    Twins\_B & $224^2$ & 56M & 300 & 80.57  & 91.25 & 79.90\\
    Twins\_B-poly & $224^2$ & 56M  & 300 & \textbf{80.59} & \textbf{100} & \textbf{80.59}   \\
    \bottomrule
  \end{tabular}
\end{table}

\subsection{Robustness tests on ImageNet 1k}
\paragraph{Settings.} We assess the robustness of the models trained from scratch in Section \ref{sec:class} under various types of transformations, using ImageNet-1K for our experiments.

\paragraph{Evaluations.} We evaluate the accuracy and consistency of the models under random cropping, horizontal flipping, random patch erasing, and random affine transformations. Additionally, we perform a worst-of-N shift attack for each batch of images, we keep the shift within a small range of $(-15, 15)$ range, to keep it inconsequential to human perception, and use the worst-case shift from the grid search. 

Some of these metrics, especially the worst-of-N shift are sensitive to the batch size used since the worst shift is chosen per batch. We use a batch size of 64 for all metrics and additionally evaluate worst-of-30 with a batch size of 1 for 2000 samples to show the lowest performance.

\paragraph{Results.} As shown in Table~\ref{tab:experiment_results_part2}, ViT\_B-poly and ViT\_S-poly obtain comparable or better accuracy than  their respective counterparts, ViT\_B and ViT\_S, under all considered transformations. Under the worst-of-k shift attack, our models achieve 100\% consistency and significantly improved accuracy, while having slight-to-high gains on the other transformations. \par 

\begin{table}[htbp]
  \centering
  \caption{Robustness experiments on ImageNet1K}
    \vspace{0.8em} 
    \label{tab:experiment_results_part2}
  \begin{tabular}{cccccc}
    \toprule
    \textbf{Model} & \textbf{Acc. Crop} & \textbf{Acc. Flip} & \textbf{Acc. Affine} & \textbf{Worst-of-30 Batch 1} & \textbf{Worst-of-30} \\
    \midrule
    ViT\_S &  75.09 & 75.50 & 69.85  &53.80 & 68.90 \\
    ViT\_S-poly  & \textbf{76.08} & \textbf{76.34} & 69.54  &\textbf{76.20} & \textbf{76.02} \\
    \midrule
    ViT\_B & 73.31 & 73.83 & 68.64 &  53.20 & 67.42 \\
    ViT\_B-poly  & \textbf{74.36} & \textbf{74.64} & \textbf{70.46} &  \textbf{74.40}& \textbf{74.20} \\ 
    \midrule
    Twins\_B & 80.51 &  80.60 & 75.88 &  62.40  & 73.86 \\
    Twins\_B-poly & 80.43 & 80.56 & \textbf{76.12} &  \textbf{80.78} &\textbf{80.78} \\
    \bottomrule
  \end{tabular}
\end{table}

\subsection{Stability and shift-equivariance tests on ImageNet-1K}
\paragraph{Settings.} We measure stability of output logits under small shifts as well as shift equivariance in the feature space using models trained from scratch in \ref{sec:class}.
\par
\begin{figure}[htbp]
  \centering
  \hfill
  \begin{subfigure}[b]{0.45\textwidth}
    \includegraphics[width=\textwidth]{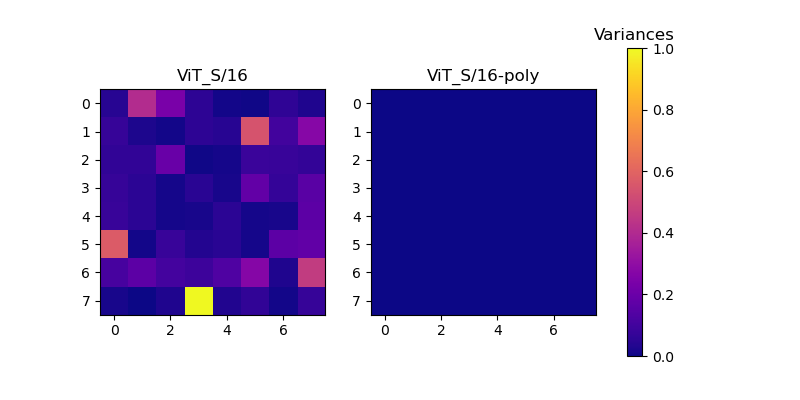}
    \label{subfig:logit-varince-vit}
    \vspace{-2em}
  \end{subfigure}
  \hfill
  \begin{subfigure}[b]{0.45\textwidth}
    \includegraphics[width=\textwidth]{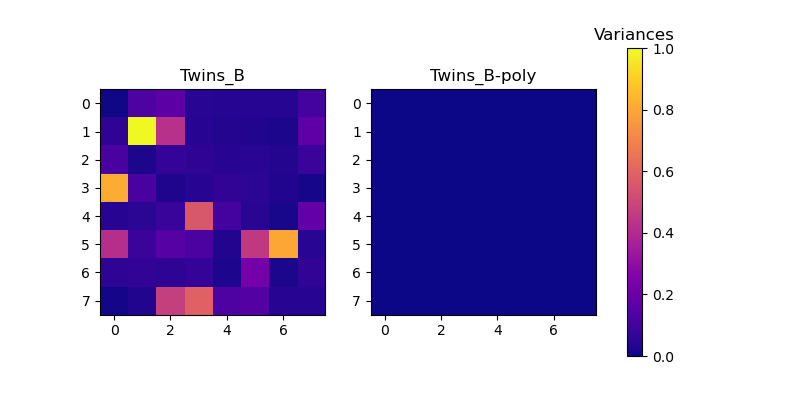}
    \label{subfig:logit-varince-twins}
    \vspace{-2em}
  \end{subfigure}\hfill
  \par
  \hfill
    \begin{subfigure}[b]{0.45\textwidth}
    \includegraphics[width=\textwidth]{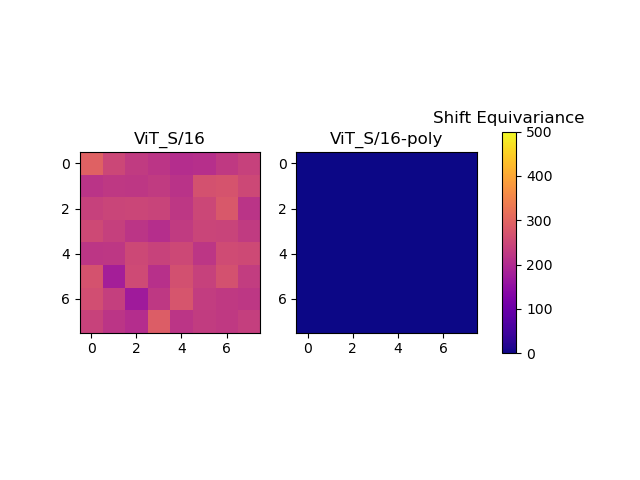}
    \label{subfig:shift-eq-vit}
    \vspace{-4em}
  \end{subfigure}
  \hfill
  \begin{subfigure}[b]{0.45\textwidth}
  \begin{center}
    \includegraphics[width=\textwidth]{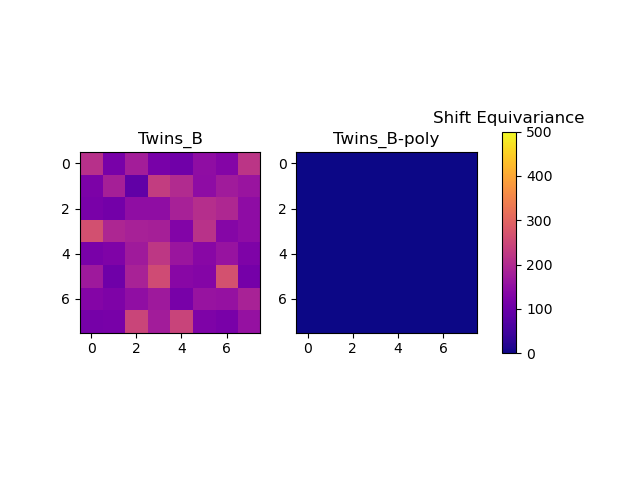}
    \label{subfig:shift-eq-twins}
    \vspace{-4em}
    \end{center}
  \end{subfigure}
  \hfill
  \caption{The top figures display the output logits variance for ViT\_S, Twins\_B, and their shift-equivariant counterparts, while the bottom figures provide a comparison of shift-equivariance tests between ViT\_S, Twins\_B, and their respective shift-equivariant versions. }
  \label{fig:logits-stability-shift-eq}
\end{figure}

 
\paragraph{Output logits variance} measures the variability of the model's logits predictions with respect to a range of small random shifts from -5 to 5.  It quantifies the spread or dispersion of the logits ($L$) as a function of the input shift. Mathematically, the output logits variance can be calculated as follows:

\begin{equation}
\text{{Variance}} = \frac{1}{N} \sum_{i=1}^{N} \left( L(x_i) - \bar{L} \right)^2
\end{equation}

where Variance represents the output logits variance, $N$ is the total number of samples, $x_i$ denotes the input sample, $L(x_i)$ corresponds to the logits prediction for the input $x_i$, $\bar{L}$ represents the mean logits prediction for the given range of input shifts,  the range of input shifts from -5 to 5 is denoted as $\Delta x$. 
As demonstrated in Figure~\ref{fig:logits-stability-shift-eq}, the output logits variance concerning small shift perturbations is almost zero for ViT\_S/16-poly and Twins\_B-poly, indicating that the output logits remain unchanged under small input shifts. Conversely, the output logits variance is nonzero for nearly 50\% of the input images, suggesting that the model's assigned probability for the input label alters in response to minor pixel shifts.

\paragraph{Shift-equivariance tests} are unit tests that measure if the features are shift-equivariant. 
Let $\mathcal{M}$ be a machine learning model that takes an input $\mathbf{X}$ and produces a feature map $\mathbf{F} = \mathcal{M}(\mathbf{X})$. For a given translation $g\in G$, define the shifted input $\mathbf{X}'$ as $\mathbf{X}' = g\cdot \mathbf{X}$. Let $\mathbf{F}' = \mathcal{M}(\mathbf{X}')$ be the feature map obtained by applying the model to the shifted input. The feature shift-equivariance test can be defined as follows:

\begin{equation}
\text{shift-equivariance}(\mathcal{M}) = \begin{cases}
    0, & \text{if } \mathbf{F}' = g'\cdot \mathbf{F} \\
    \|\mathbf{F}' -\mathbf{F} \| , & \text{otherwise}
\end{cases}
\end{equation}
where $g'\in G$ is translation in the feature space, $\|\cdot \|$ is $L_2$ norm. As demonstrated in Figure~\ref{fig:logits-stability-shift-eq}, both the polyphase models of ViT and Twins successfully pass all shift equivariance tests in the feature space, while the original ViT and Twins models fail to do so and exhibit substantial norm differences in the feature space.

\section{Related Work}

\paragraph{Data augmentation} encourages shift-equivariance by adding shifted copies of images to the training set but lacks guarantees. In CNNs, it has been shown that models learn invariance to transformations only for images similar to typical training set images \citep{poorgen}.\par

\paragraph{Regularization during training} encourages shift-equivariance and invariance by imposing soft constraints. A pretraining objective during self-supervised learning can be added to predict transformations applied to the input \citep{essl}. A loss function based on cross-correlation of embedded features encourages equivariance \citep{embedloss-equiv}. However, these regularizations do not guarantee shift-equivariance.\par 

\paragraph{Architectural design} can also result in shift-equivariance and invariance. For CNNs, anti-aliasing strategies \citep{blurpool} and adaptive polyphase sampling (APS) address the lack of shift-equivariance due to downsampling. The former lacks guarantees, while the latter is computationally expensive, requiring full convolution computations that can be problematic when applied to subsampled attentions. Subsampled attentions address the computational complexity of global attention; computing global attention before subsampling not only increases computation but also defeats the purpose of subsampled attentions. 

\section{Conclusions and Discussions}
In this paper, we revive shift equivariance in vision transformers by presenting versatile and adaptable modules that seamlessly integrate with vision transformer models, leading to improved performance. We introduce an input-adaptive nonlinear operator that guarantees shift-equivariance, resulting in more robust and dependable outcomes across a range of visual transformations. We provide theoretical guarantees of shift equivariance for strided convolution, window attention, and GSA that utilize polyphase anchoring. Through carefully controlled experiments, we demonstrate that vision transformers with polyphase anchoring and depthwise convolution achieve 100\% consistency in classification, obtain an average of 20\% percentage point gain (or 37\% increase) under the worst-of-30 shift attack, and reach comparable or superior performance under cropping, horizontal flipping, and affine transformations.


Due to the high computational cost of training from scratch, this paper, conducted in an academic setup, focuses on conducting controlled experiments for comparative analysis rather than optimizing for higher accuracy. Although the performance of the models may not achieve state-of-the-art standards, the primary goal of this paper is to investigate whether the geometric guarantees hold in practice and whether shift-equivariance results in more robust models.
We defer the work of using industrial-scale computing resources to obtaining state-of-the-art performance to the future. 

\newpage
\bibliographystyle{plainnat}
\bibliography{references}

\begin{thebibliography}{17}
\providecommand{\natexlab}[1]{#1}
\providecommand{\url}[1]{\texttt{#1}}
\expandafter\ifx\csname urlstyle\endcsname\relax
  \providecommand{\doi}[1]{doi: #1}\else
  \providecommand{\doi}{doi: \begingroup \urlstyle{rm}\Url}\fi

\bibitem[Azulay and Weiss(2018)]{poorgen}
Aharon Azulay and Yair Weiss.
\newblock Why do deep convolutional networks generalize so poorly to small
  image transformations?
\newblock \emph{CoRR}, abs/1805.12177, 2018.
\newblock URL \url{http://arxiv.org/abs/1805.12177}.

\bibitem[Chaman and Dokmanic(2021)]{poly-d}
Anadi Chaman and Ivan Dokmanic.
\newblock Truly shift-invariant convolutional neural networks.
\newblock In \emph{Proceedings of the IEEE/CVF Conference on Computer Vision
  and Pattern Recognition (CVPR)}, pages 3773--3783, June 2021.

\bibitem[Chu et~al.(2021)Chu, Tian, Wang, Zhang, Ren, Wei, Xia, and
  Shen]{twins}
Xiangxiang Chu, Zhi Tian, Yuqing Wang, Bo~Zhang, Haibing Ren, Xiaolin Wei,
  Huaxia Xia, and Chunhua Shen.
\newblock Twins: Revisiting the design of spatial attention in vision
  transformers.
\newblock In A.~Beygelzimer, Y.~Dauphin, P.~Liang, and J.~Wortman Vaughan,
  editors, \emph{Advances in Neural Information Processing Systems}, 2021.
\newblock URL \url{https://openreview.net/forum?id=5kTlVBkzSRx}.

\bibitem[Chu et~al.(2023)Chu, Tian, Zhang, Wang, and Shen]{conditional}
Xiangxiang Chu, Zhi Tian, Bo~Zhang, Xinlong Wang, and Chunhua Shen.
\newblock Conditional positional encodings for vision transformers, 2023.

\bibitem[Dai et~al.(2021)Dai, Liu, Le, and Tan]{coatnet}
Zihang Dai, Hanxiao Liu, Quoc~V. Le, and Mingxing Tan.
\newblock Coatnet: Marrying convolution and attention for all data sizes, 2021.

\bibitem[Dangovski et~al.(2022)Dangovski, Jing, Loh, Han, Srivastava, Cheung,
  Agrawal, and Soljacic]{essl}
Rumen Dangovski, Li~Jing, Charlotte Loh, Seungwook Han, Akash Srivastava, Brian
  Cheung, Pulkit Agrawal, and Marin Soljacic.
\newblock Equivariant self-supervised learning: Encouraging equivariance in
  representations.
\newblock In \emph{International Conference on Learning Representations}, 2022.
\newblock URL \url{https://openreview.net/forum?id=gKLAAfiytI}.

\bibitem[Ding et~al.(2022)Ding, Xiao, Codella, Luo, Wang, and Yuan]{davit}
Mingyu Ding, Bin Xiao, Noel Codella, Ping Luo, Jingdong Wang, and Lu~Yuan.
\newblock Davit: Dual attention vision transformers.
\newblock In \emph{Computer Vision--ECCV 2022: 17th European Conference, Tel
  Aviv, Israel, October 23--27, 2022, Proceedings, Part XXIV}, pages 74--92.
  Springer, 2022.

\bibitem[Dong et~al.(2021)Dong, Bao, Chen, Zhang, Yu, Yuan, Chen, and
  Guo]{cswin}
Xiaoyi Dong, Jianmin Bao, Dongdong Chen, Weiming Zhang, Nenghai Yu, Lu~Yuan,
  Dong Chen, and Baining Guo.
\newblock Cswin transformer: A general vision transformer backbone with
  cross-shaped windows, 2021.

\bibitem[Dosovitskiy et~al.(2020)Dosovitskiy, Beyer, Kolesnikov, Weissenborn,
  Zhai, Unterthiner, Dehghani, Minderer, Heigold, Gelly, et~al.]{vit}
Alexey Dosovitskiy, Lucas Beyer, Alexander Kolesnikov, Dirk Weissenborn,
  Xiaohua Zhai, Thomas Unterthiner, Mostafa Dehghani, Matthias Minderer, Georg
  Heigold, Sylvain Gelly, et~al.
\newblock An image is worth 16x16 words: Transformers for image recognition at
  scale.
\newblock In \emph{International Conference on Learning Representations}, 2020.

\bibitem[Gruver et~al.(2022)Gruver, Finzi, Goldblum, and Wilson]{lie}
Nate Gruver, Marc Finzi, Micah Goldblum, and Andrew~Gordon Wilson.
\newblock The lie derivative for measuring learned equivariance, 2022.

\bibitem[Liu et~al.(2021)Liu, Lin, Cao, Hu, Wei, Zhang, Lin, and Guo]{swin}
Ze~Liu, Yutong Lin, Yue Cao, Han Hu, Yixuan Wei, Zheng Zhang, Stephen Lin, and
  Baining Guo.
\newblock Swin transformer: Hierarchical vision transformer using shifted
  windows.
\newblock In \emph{Proceedings of the IEEE/CVF International Conference on
  Computer Vision (ICCV)}, 2021.

\bibitem[Liu et~al.(2022)Liu, Hu, Lin, Yao, Xie, Wei, Ning, Cao, Zhang, Dong,
  Wei, and Guo]{swinv2}
Ze~Liu, Han Hu, Yutong Lin, Zhuliang Yao, Zhenda Xie, Yixuan Wei, Jia Ning, Yue
  Cao, Zheng Zhang, Li~Dong, Furu Wei, and Baining Guo.
\newblock Swin transformer v2: Scaling up capacity and resolution, 2022.

\bibitem[Tu et~al.(2022)Tu, Talebi, Zhang, Yang, Milanfar, Bovik, and
  Li]{maxvit}
Zhengzhong Tu, Hossein Talebi, Han Zhang, Feng Yang, Peyman Milanfar, Alan
  Bovik, and Yinxiao Li.
\newblock Maxvit: Multi-axis vision transformer.
\newblock \emph{ECCV}, 2022.

\bibitem[Wang et~al.(2021)Wang, Xie, Li, Fan, Song, Liang, Lu, Luo, and
  Shao]{pvt}
Wenhai Wang, Enze Xie, Xiang Li, Deng-Ping Fan, Kaitao Song, Ding Liang, Tong
  Lu, Ping Luo, and Ling Shao.
\newblock Pyramid vision transformer: A versatile backbone for dense prediction
  without convolutions.
\newblock In \emph{Proceedings of the IEEE/CVF International Conference on
  Computer Vision}, pages 568--578, 2021.

\bibitem[Xie et~al.(2022)Xie, Wen, Lau, Ur~Rehman, and Shen]{embedloss-equiv}
Yuyang Xie, Jianhong Wen, Kin~Wai Lau, Yasar~Abbas Ur~Rehman, and Jiajun Shen.
\newblock What should be equivariant in self-supervised learning.
\newblock In \emph{2022 IEEE/CVF Conference on Computer Vision and Pattern
  Recognition Workshops (CVPRW)}, pages 4110--4119, 2022.
\newblock \doi{10.1109/CVPRW56347.2022.00456}.

\bibitem[Zhang et~al.(2021)Zhang, Dai, Yang, Xiao, Yuan, Zhang, and
  Gao]{longformer}
Pengchuan Zhang, Xiyang Dai, Jianwei Yang, Bin Xiao, Lu~Yuan, Lei Zhang, and
  Jianfeng Gao.
\newblock Multi-scale vision longformer: A new vision transformer for
  high-resolution image encoding.
\newblock In \emph{Proceedings of the IEEE/CVF International Conference on
  Computer Vision}, pages 2998--3008, 2021.

\bibitem[Zhang(2019)]{blurpool}
Richard Zhang.
\newblock Making convolutional networks shift-invariant again, 2019.

\end{thebibliography}

\end{document}


\section*{A. Supplementary Material}
\subsection*{A.1 Positional Encoding}
In section 3.1 of the main paper, we mentioned that relative positional encoding \citep{swin, swinv2}  is not shift-equivariant. We reiterate the definition below and provide a counterexample. Relative positional encoding is defined as: 
\begin{align}
     A_r = \text{SoftMax}(XW_Q (XW_K)^T+B )XW_V
\end{align}
Counterexample:
Let $B$ be a $n\times n $ square matrix with two standard basis vectors $e_1$ and $e_2$ and everywhere else zero. 
\begin{align}
    B = \left (
    \begin{array}{cccc}
        1 & 0 & 0  & \dots  \\
        0 & 1 & 0 & \dots \\
        0 & 0 & 0 & \dots \\
        \vdots & \vdots & \vdots & \vdots 
    \end{array}
    \right )
\end{align}
Let $P_{\pi}$ be the matrix representation for the linear transformation $T_{\pi}$ that circularly shifts the input signals s.t
\begin{align}
    P = \left (
    \begin{array}{c}
        (e_n)^T \\ 
        (e_1)^T \\ 
        (e_2)^T \\
        \vdots \\ 
        (e_{n-1})^T
    \end{array}
    \right ).
\end{align}
For relative positional encoding to be shift-equivariant, we must have $ A_r (T_{\pi}(X))=  T_{\pi}( A_r(X))$. 

\begin{align}
    LHS = A_r (T_{\pi}(X)) &= \text{SoftMax}(T_{\pi}(X)W_Q (T_{\pi}(X)W_K)^T+B )T_{\pi}(X)W_V \\
    &= \text{SoftMax}(P_{\pi}XW_Q (P_{\pi}XW_K)^T+B )P_{\pi}XW_V 
\end{align}

\begin{align}
    RHS = T_{\pi}( A_r(X)) &= P_{\pi} \text{SoftMax}(XW_Q (XW_K)^T) + B)P_{\pi}^TP_{\pi}XW_V  \\ 
    &= \text{SoftMax} (P_{\pi} XW_Q (XW_K)^T)P_{\pi}^T + P_{\pi}BP_{\pi}^T)P_{\pi}XW_V 
\end{align}
Assume  $P_{\pi}XW_V$ is right-invertible: $\exists Q$ s.t  $(P_{\pi}XW_V) Q = I$. Multiply both LHS and RHS by $Q$ and apply logarithmic function. 
\begin{align}
    LHS  =  P_{\pi} XW_Q (XW_K)^T)P_{\pi}^T + B+ \log(S_1) 
\end{align}
\begin{align}
    RHS = P_{\pi} XW_Q (XW_K)^T)P_{\pi}^T + P_{\pi}BP_{\pi}^T + \log(S_2) 
\end{align}
For $LHS = RHS$, the following much hold:
\begin{align}
    B = P_{\pi}BP_{\pi}^T + C,
\end{align}
where $C$ is a constant matrix. However,
\begin{align}
    P_{\pi}BP_{\pi}^T = \left (\begin{array}{cccc}
        0 &  e_2 & e_3 & \dots \\
    \end{array} \right ) 
\end{align}

QED.\par 

Tangent from the solutions proposed in the main paper, we reveal that relative positional encoding is shift equivariant under specific conditions. More concretely, if the bias term is shift equivariant, relative positional encoding is shift equivariant \citep{swin,swinv2}. Let $T_{\pi}$ denote the spatial translation of the input $X$, and $A_r$ denote a self-attention operator with relative position bias. We have: 
\begin{align}
    A_r (T_{\pi}(X)) &= \text{SoftMax}(T_{\pi}(X)W_Q (T_{\pi}(X)W_K)^T+B )T_{\pi}(X)W_V \\
    &= \text{SoftMax}(P_{\pi}XW_Q (P_{\pi}XW_K)^T+B )P_{\pi}XW_V \\ 
    &= \text{SoftMax} (P_{\pi} XW_Q (XW_K)^TP_{\pi}^T + P_{\pi}P_{\pi}^TB)P_{\pi}XW_V \label{eq:lhs}\\
    &= \text{SoftMax} (P_{\pi} XW_Q (XW_K)^TP_{\pi}^T + P_{\pi}BP_{\pi}^T)P_{\pi}XW_V \label{eq:rhs} \\    
    &= P_{\pi} \text{SoftMax}(XW_Q (XW_K)^T) + B)P_{\pi}^TP_{\pi}XW_V \\
    &=T_{\pi}( A_r(X))
\end{align}

Although it is not directly related to the solutions proposed in the main manuscript, this finding demonstrates that shift-equivariance can be ensured in relative positional encoding through constraining the bias term to be shift-equivariant. \par 

\subsection*{A.2 Polyphase anchoring}
In section 3.2 of the main manuscript, we claimed that the composition of polyphase anchoring with strided convolution, window attention, and global subsampled attention respectively results in shift-equivariant operations. We provide proofs for those claims in this section. \par 

\begin{lemma}
    \label{lemma:shift}
    Polyphase anchoring operator $P$ is general equivariant with respect to $\forall g\in G$, where G is the symmetry group of translations, and $P: V \to V$ is a nonlinear operator that conditionally shift the input $X$. $\forall g\in G$, $\exists g' \in G$ s.t:
    \begin{equation}
    P(g \cdot X) = g' \cdot P(X),
    \end{equation}
    where $\cdot$ denotes the linear mapping of the input by the representation of group elements in $G$. \par 
\end{lemma}
Proof: 
let $X \in \mathbb{R}^{\cdots \times H \times W }$,
\begin{align}
    P(g\cdot X) = g_{\mid (g\cdot X)} \cdot g \cdot X
\end{align}
where $ g_{\mid (g\cdot X)}$ is some translation conditioned on input $g\cdot X$.
\begin{align}
    P(X) = g_{\mid X} \cdot X 
\end{align}
where $ g_{\mid X}$ is some translation conditioned on input $X$. 
Since $g_{\mid X}, g_{\mid (g\cdot X)}, g \in G$, $\exists g' \in G$ s.t 
\begin{align}
    g_{\mid (g\cdot X)} \cdot g \cdot X = g' \cdot g_{\mid X} \cdot X\\
    P(g\cdot X) = g'\cdot P(X)
\end{align}
QED.\par 
\begin{corollary}
\label{coro:1}
$P(g\cdot X) = g'\cdot P(X)$ where $g'$ translates $ P(X)$ by an integer multiple of stride size $s$. Stride size is the distance between two consecutive tokens in the same polyphase on a 2D grid. 
\end{corollary}
Proof: let $X \in \mathbb{R}^{\cdots \times H \times W}$, $X[:,i,j]  \in \mathbb{C} $ denote a token located at $(i,j)$ coordinate on a 2D grid. \par 
By definition of polyphase anchoring, tokens in the maximum polyphase are at the anchor positions s.t 
\begin{align}
    P(X)[:,0::s,0::s] = \argmax_{P(X)[:,i::s,j::s] \in \{P(X)[:,i::s,j::s] | i,j\in \mathbb{Z}, i,j < s \}}\|P(X)[:,i::s,j::s] \|, 
\end{align}
where $P(X)[:,0::s,0::s]$ denotes the polyphase or subsampled grid starting from top left at $(0,0)$ with stride size $s$. (This notation aligns with regular PyTorch usage.)  Assuming that maximum polyphase is unique, $P(X)[:,0,0]$ $P(g\cdot X)[:,0,0]$ both belong to the same polyphase. Since coordinate distance between tokens in the same polyphase is a integer multiple of stride size, we must have $P(g\cdot X) = g'\cdot P(X)$, where $g'$ translate $P(X)$ by a multiple of stride size $s$ on a 2D grid. QED. \par 

\begin{lemma}
      Given a window attention operator $A_w$ and polyphase anchoring operator $P$, the composition of these operators is general shift-equivariant $\forall g \in G$, where $G$ is the translation group, and for $s, w \in \mathbb{Z}$ such that $s = w$, where $s$ is the stride size in the polyphase and $w \times w$ is the window size. This can be expressed as:
     \begin{equation}A_w (P(g\cdot X)) = g' \cdot A_w (P(X))\end{equation}
\end{lemma}
Proof: 
let $X \in \mathbb{R}^{C\times H \times W}$,
$X=\left[\begin{array}{ccc}
   X_{00}  & \cdots & X_{0n} \\
    \vdots &  \vdots & \vdots \\
    X_{m0} & \cdots  & X_{mn}
\end{array} \right ]$
where $m = \frac{H}{w}$, $n = \frac{W}{w}$, and $X_{ij} \in R^{C\times w \times w}$, $i \in \{0,\cdots,m \}, j \in \{0,\cdots, n\}$. We call $w\times w$ window size and $X_{ij}$ tokens in the window $(i,j)$. The window attention operator 
$$A_w(X) = 
\left [ \begin{array}{ccc}
    A_s(X_{00})  & \cdots & A_s(X_{0n}) \\
    \vdots &  \vdots & \vdots \\
    A_s(X_{m0}) & \cdots  & A_s(X_{mn})
\end{array} \right ]
$$, 
where $A_s$ is the self attention operator $$A_s(X) = \text{SoftMax}(XW_q (XW_k)^T)XW_v = \text{SoftMax}(Q (K)^T)V $$.
$$A_w(P(X)) = A_w \left (
\left [ \begin{array}{ccc}
    \hat{X}_{00}  & \cdots & \hat{X}_{0n} \\
    \vdots &  \vdots & \vdots \\
    \hat{X}_{m0} & \cdots  & \hat{X}_{mn}
\end{array} \right ] \right )
$$, where $\{\hat{X}_{00}[:,0,0], \cdots, \hat{X}_{mn}[:,0,0] \}$ are tokens in the maximum polyphase because the polyphase anchoring algorithm conditionally shifts the input data so that the maximum polyphase is $\hat{X}[:,0::w, 0::w]$. 
\begin{equation}
\begin{split}
    A_w (P(g\cdot X)) &= A_w (g'\cdot P(X)) \\ 
    &= A_w \left ( 
    g' \cdot \left [ \begin{array}{ccc}
    \hat{X}_{00}  & \cdots & \hat{X}_{0n} \\
    \vdots &  \vdots & \vdots \\
    \hat{X}_{m0} & \cdots  & \hat{X}_{mn}
\end{array} \right ] \right )\\
&= 
 \left ( 
   \left [ \begin{array}{ccc}
    g' \cdot A_s(\hat{X}_{ij})  & \cdots &  g' \cdot A_s(\hat{X}_{i(j-1)}) \\
    \vdots &  \vdots & \vdots \\
   g' \cdot  A_s(\hat{X}_{(i-1)j}) & \cdots  & g' \cdot A_s(\hat{X}_{(i-1)(j-1)})
\end{array} \right ] \right )  \quad \text{(Corollary \ref{coro:1})} \\ 
&= 
g' \cdot
\left ( 
   \left [ \begin{array}{ccc}
    A_s(\hat{X}_{00})  & \cdots & A_s(\hat{X}_{0n}) \\
    \vdots &  \vdots & \vdots \\
    A_s(\hat{X}_{m0}) & \cdots  & A_s(\hat{X}_{mn})
\end{array} \right ] \right )  \\
&= g' \cdot A_w(P(X))
\end{split}
\end{equation}
QED.\par 



\begin{lemma}
      Let $P$ be the polyphase anchoring operator and $\ast_s$ represent the strided convolution operator. $\forall g \in G$, where $G$ is the translation group, and $\forall s_1, s_2 \in \mathbb{Z}$ s.t $s_1 = s_2$, where $s_1$ is the stride size in the polyphase  and $s_2$ is the stride size of convolution, the composition of strided convolution and polyphase anchoring is general shift-equivariant:
     \begin{align}
           h \ast_s P(g\cdot X ) =  g' \cdot (h \ast_s  P( X ))
     \end{align}
\end{lemma}
 Here, $X$ denotes the input signal, $h$ is the convolution filter, and $\cdot$ signifies the linear mapping of the input by the representation of group elements in $G$. Furthermore, $P: V \to V$ is a nonlinear operator acting on the input space $V$.
 Mathematically, strided convolution $\ast_s$ can be represented as a full convolution followed by a downsampling operation
 \begin{align*}
      h \ast_s X  =  P^{(s)}_{0,0} (h \ast X )
 \end{align*}
 where $P^{(s)}_{m,n}(\cdot)$ is a function with a matrix as input and its down-sampled sub-matrix as output. This function select the elements on the grid defined by $m, n, s$, where $(m, n)$ denotes the upperleft position of the grid, and $s$ denotes the sub-sampling stride. \par 
\begin{align}
    LHS =  h \ast_s P(g\cdot X ) &=  P^{(s)}_{0,0}(h \ast P(g\cdot X ))  \\
    &=  P^{(s)}_{0,0}(h \ast (g'\cdot P( X ))) \\ 
    & = P^{(s)}_{0,0}(g'\cdot (h \ast P( X )) ) \label{pg}\\
    & = g'\cdot (P^{(s)}_{0,0}( h \ast P( X ) ) )\label{gp} \\ 
    & = RHS
\end{align}
where $g'\in G$ translates input by an integer multiple of stride size $s$.  QED.\par 

\begin{lemma}
    For a global subsampled attention operator $A_g$ \citep{twins} combined with a polyphase anchoring operator $P$, general shift-equivariance is achieved for $\forall g \in G$, where $G$ is the translation group, and for $s_1, s_2 \in \mathbb{Z}$ such that $s_1 = s_2$. Here, $s_1$ is the stride size in the polyphase, and $s_2$ is the stride size in the global subsampled attention. This can be expressed as:
     \begin{equation}
     A_g (P(g\cdot X)) = g' \cdot A_g (P(X))
     \end{equation}
\end{lemma}
    
In global subsampled attention, we have: \begin{equation} 
A_g(X) = \text{SoftMax}(Q K_{s}^T)V_{s} ,
\end{equation}
where $K_s$ and $V_s$ are subsampled from the full keys $K$ and values $V$ using strided convolution. \par

\begin{align}
    LHS &= A_g (P(g\cdot X)) \\
    &= \text{SoftMax}(P(g\cdot X)W_q ( h\ast_s P(g\cdot X) W_k )^T) h\ast_s P(g\cdot X) W_v \\
    & = \text{SoftMax}(g'\cdot P(X)W_q ( h\ast_s (g'\cdot P(X)) W_k )^T) h\ast_s (g'\cdot P( X)) W_v  \\ 
    & = \text{SoftMax}(g'\cdot P(X)W_q ( g'\cdot  (h\ast_s P(X) )W_k )^T) g'\cdot  (h\ast_s P(X) )W_v \\ 
    & = \text{SoftMax}(P_{g'}  P(X)W_q ( P_{g'} (h\ast_s P(X) )W_k )^T) P_{g'}  (h\ast_s P(X) )W_v \\ 
    & = \text{SoftMax}(P_{g'}  P(X)W_q  ( (h\ast_s P(X) )W_k )^T P_{g'}^T) P_{g'} (h\ast_s P(X) )W_v  \\ 
    & = P_{g'}  \text{SoftMax}( P(X)W_q  ( (h\ast_s P(X) )W_k )^T ) P_{g'}^T P_{g'} (h\ast_s P(X) )W_v   \\ 
    & = P_{g'}  \text{SoftMax}( P(X)W_q  ( (h\ast_s P(X) )W_k )^T ) (h\ast_s P(X) )W_v  \\
    &= g' \cdot A_g(P(X)) = RHS,
\end{align}
where $P_{g'}$ is the matrix representation of a group element $g'\in G$ from the symmetry group of translations. QED.\par

\subsection*{A.3 Composition of equivariant functions }
In Section 3 of the main manuscript, we conduct a comprehensive analysis of Vision Transformers (ViT) and their derivatives, focusing on the aspect of shift-equivariance. We identify specific modules within these models that do not preserve shift-equivariance, which is integral to maintaining spatial coherence in vision tasks. In response to this discovery, we propose and implement a series of corrective measures, facilitating the design of fully shift-equivariant Vision Transformer architectures. Importantly, our approach capitalizes on the property that a composite function constructed from shift-equivariant functions retains shift-equivariance. This results in models that preserve spatial information across the entire network architecture. 
\begin{lemma}
 Composition of two equivariant functions with respect to transformations in symmetry group is equivariant. 
\end{lemma}
Proof: Let $G$ be a symmetry group and let $f:X\to Y$ and $h:Y\to Z$ be equivariant functions, i.e., for all $x\in X$ and $g\in G$, we have:
$$
f(g\cdot x) = g\cdot f(x)
$$
and
$$
h(g'\cdot y) = g'\cdot h(y)
$$
where $\cdot$ denotes the group action of $G$ on $X$, $Y$ and $Z$. We want to show that $h\circ f: X\to Z$ is also equivariant, i.e., for all $x\in X$ and $g\in G$, we have:

$$
(h\circ f)(g\cdot x) = g\cdot (h\circ f)(x)
$$

We start with the left-hand side:

$$
\begin{aligned}
(h\circ f)(g\cdot x) &= h(f(g\cdot x)) && \text{(definition of composition)} \\
&= h(g\cdot f(x)) && \text{(by equivariance of $f$)} \\
&= g\cdot h(f(x)) && \text{(by equivariance of $h$)} \\
&= g\cdot (h\circ f)(x) && \text{(definition of composition)}
\end{aligned}
$$

where we used $g'\cdot h(y) = h(g'\cdot y)$, the associative property of group action, and the equivariance of $f$ and $h$.

Therefore, we have shown that $(h\circ f)(g\cdot x) = g\cdot (h\circ f)(x)$ for all $x\in X$ and $g\in G$, which means that $h\circ f$ is equivariant with respect to the group action of $G$.\par

\newpage
\bibliographystyle{plainnat}
\bibliography{references}